\documentclass[preprint,review,12pt,numbers]{elsarticle}

\usepackage[utf8]{inputenc}
\usepackage[T1]{fontenc}
\usepackage{lmodern}
\usepackage{amsmath,amssymb,amsfonts}
\usepackage{booktabs}
\usepackage{graphicx}
\usepackage{hyperref}
\usepackage{xcolor}
\usepackage{multirow}
\usepackage{array}
\usepackage{algorithm}
\usepackage{algorithmic}
\usepackage{microtype}
\usepackage{float}
\usepackage{url}
\graphicspath{{figures/}}

\hypersetup{
    colorlinks=true,
    linkcolor=blue!60!black,
    citecolor=blue!60!black,
    urlcolor=blue!60!black
}

\begin{document}

\begin{frontmatter}

\title{Component-Aware Self-Speculative Decoding in Hybrid Language Models}

\author[inst1]{Hector Borobia\fnref{cor1}}
\ead{hecboar@doctor.upv.es}
\author[inst2]{Elies Segu\'i-Mas}
\ead{eseguim@upvnet.upv.es}
\author[inst3]{Guillermina Tormo-Carb\'o}
\ead{gtormo@omp.upv.es}

\fntext[cor1]{Corresponding author.}

\affiliation[inst1]{organization={VRAIN -- Valencian Research Institute for Artificial Intelligence, Universitat Polit\`ecnica de Val\`encia},
    city={Valencia},
    country={Spain}}

\affiliation[inst2]{organization={Department of Economics and Social Sciences, Universitat Polit\`ecnica de Val\`encia},
    city={Valencia},
    country={Spain}}

\affiliation[inst3]{organization={Department of Business Organisation, Universitat Polit\`ecnica de Val\`encia},
    city={Valencia},
    country={Spain}}

\begin{abstract}
Speculative decoding accelerates autoregressive inference by drafting candidate tokens with a fast model and verifying them in parallel with the target. Self-speculative methods avoid the need for an external drafter but have been studied exclusively in homogeneous Transformer architectures. We introduce \emph{component-aware self-speculative decoding}, the first method to exploit the internal architectural heterogeneity of hybrid language models, isolating the SSM/linear-attention subgraph as a zero-cost internal draft. We evaluate this on two architecturally distinct hybrid families: Falcon-H1 (parallel: Mamba-2 + attention per layer) and Qwen3.5 (sequential: interleaved linear and attention layers), with a pure Transformer control (Qwen2.5). Parallel hybrids achieve acceptance rates of $\alpha = 0.68$ at draft length $k=2$ under greedy decoding, while sequential hybrids yield only $\alpha = 0.038$---an $18\times$ gap attributable to how each architecture integrates its components. The property is scale-invariant: Falcon-H1 at 3B reproduces the rates observed at 0.5B. We further show that perplexity degradation from a companion ablation study predicts speculative viability without running speculative decoding: a $3.15\times$ ratio (Falcon) maps to $\alpha = 0.37$ at $k=4$, while $81.96\times$ (Qwen) maps to $\alpha = 0.019$. For sequential hybrids, generic LayerSkip achieves $12\times$ higher acceptance rates than the component-aware strategy. The composition pattern of hybrid models---not merely the presence of alternative components---determines whether component-level self-speculation is viable.
\end{abstract}


\end{frontmatter}

\section{Introduction}
\label{sec:introduction}

Autoregressive decoding remains the primary bottleneck for inference latency in large language models (LLMs). Each token generation requires a full forward pass through the model, making generation time proportional to sequence length regardless of hardware parallelism. Speculative decoding~\cite{leviathan2023fast,chen2023accelerating} addresses this by using a fast \emph{draft model} to propose multiple candidate tokens, which the target model then verifies in a single parallel forward pass. When draft tokens are accepted, multiple tokens are produced per step, yielding wall-clock speedups of 1.5--$3\times$ while preserving the target model's output distribution exactly~\cite{leviathan2023fast}.

A key practical challenge is obtaining a suitable draft model. External drafters require separate model storage and careful alignment with the target~\cite{choi2025mamba}, while self-speculative methods---which derive the drafter from the target model itself---avoid this overhead entirely. LayerSkip~\cite{elhoushi2024layerskip} constructs an internal draft by exiting early; CLaSp~\cite{chen2025clasp} dynamically selects layers to skip via dynamic programming; SWIFT~\cite{xia2024swift} adaptively identifies skippable layers on the fly. All three achieve 1.3--$2.2\times$ speedups on homogeneous Transformer architectures by exploiting layer-level redundancy.

A parallel development in model architecture has been the emergence of \emph{hybrid language models} that combine softmax attention with alternative sequence-processing components: state space models (SSMs) such as Mamba-2~\cite{dao2024transformers} or linear attention mechanisms like Gated DeltaNet~\cite{yang2024gated}. The Falcon-H1 family~\cite{zuo2025falcon} employs a \emph{parallel} design where each block processes its input through both an SSM and an attention head simultaneously, summing their outputs. Qwen3.5~\cite{qwen2025qwen3} adopts a \emph{sequential} design, interleaving dedicated linear attention layers with full softmax attention layers. Other notable hybrids include Jamba~\cite{lieber2024jamba}, which combines Transformer and Mamba layers with mixture-of-experts.

These hybrid architectures present a unique opportunity for self-speculative decoding that existing methods cannot exploit. Rather than skipping layers (which treats the model as a homogeneous stack), one can \emph{isolate the alternative component subgraph}---running only the SSM or linear attention pathway and suppressing attention---to produce a computationally cheaper draft. The SSM pathway operates in $O(1)$ memory per token (via recurrent state, with no KV cache), providing a natural computational advantage for drafting.

However, whether this \emph{component-aware self-speculation} is viable depends on how close the SSM-only output distribution is to the full hybrid model's distribution. If the attention pathway contributes only refinements, the SSM subgraph should produce good draft tokens. If attention is essential for core language modeling, the SSM-only distribution will diverge catastrophically, and most draft tokens will be rejected.

In this paper, we investigate this question systematically. We introduce component-aware self-speculative decoding for hybrid LLMs and evaluate it across two fundamentally different hybrid architectures. Our contributions are:

\begin{enumerate}
    \item We propose \emph{component-aware self-speculative decoding}, the first self-speculation method that exploits architectural heterogeneity in hybrid models by isolating the SSM/linear-attention subgraph as a zero-cost internal draft model (Section~\ref{sec:method}).

    \item We demonstrate a stark \emph{architectural determinism}: parallel hybrids (Falcon-H1) achieve acceptance rates of $\alpha = 0.68$ at $k=2$, while sequential hybrids (Qwen3.5) yield only $\alpha = 0.038$, establishing that the component integration pattern---not merely the presence of alternative components---determines speculative viability (Section~\ref{sec:results_arch}).

    \item We show this property is \emph{scale-invariant}: Falcon-H1 at 3B parameters reproduces the acceptance rates observed at 0.5B, ruling out model size as a confound (Section~\ref{sec:results_scale}).

    \item We establish that \emph{functional component ablation predicts speculative viability}: the perplexity degradation ratio from a companion study~\cite{borobia2026ablation} exhibits a perfect inverse correlation with acceptance rate, providing a cheap diagnostic for whether component-aware self-speculation will work on a given architecture (Section~\ref{sec:results_correlation}).

    \item We provide a \emph{strategy comparison} showing that for sequential hybrids where component-aware self-speculation fails, generic LayerSkip achieves $12\times$ higher acceptance rates, while for parallel hybrids, component-aware self-speculation remains the best available strategy (Section~\ref{sec:results_strategies}).
\end{enumerate}

\section{Background and Related Work}
\label{sec:related}

\subsection{Speculative Decoding}

Speculative decoding was independently proposed by Leviathan et al.~\cite{leviathan2023fast} and Chen et al.~\cite{chen2023accelerating}. The core idea is a draft-then-verify paradigm: a small, fast model generates $k$ candidate tokens autoregressively, and the target model verifies all $k$ tokens in a single forward pass. Accepted tokens are guaranteed to follow the target model's distribution exactly through a rejection sampling scheme. When the draft model closely approximates the target, multiple tokens are accepted per verification step, yielding wall-clock speedups proportional to the expected number of accepted tokens.

The theoretical speedup depends on two factors: the acceptance rate $\alpha$ (the probability that a draft token matches the target distribution sufficiently to be accepted) and the computational cost ratio between drafting and verification. Formally, for $k$ draft tokens with per-token acceptance probability $\alpha$, the expected number of tokens generated per round is $(1 - \alpha^{k+1})/(1 - \alpha)$, and the speedup is approximately:
\begin{equation}
S \approx \frac{(1 - \alpha^{k+1})/(1 - \alpha)}{1 + k \cdot c_{\text{draft}}/c_{\text{verify}}}
\label{eq:speedup}
\end{equation}
where $c_{\text{draft}}/c_{\text{verify}}$ is the FLOP ratio of the draft model to the target model.

\subsection{Self-Speculative Decoding}

Self-speculative methods eliminate the need for a separate draft model by deriving the drafter from the target itself. LayerSkip~\cite{elhoushi2024layerskip} trains with progressive layer dropout and shared early exits, enabling self-speculative decoding where early layers draft tokens and remaining layers verify. It achieves speedups of up to $2.16\times$ on LLaMA models. CLaSp~\cite{chen2025clasp} introduces a plug-and-play layer-skipping strategy using dynamic programming to optimize which layers to skip, achieving 1.3--$1.7\times$ speedups on LLaMA-3 without additional training. SWIFT~\cite{xia2024swift} adaptively selects layers to skip based on input-dependent layer sparsity, reaching 1.3--$1.6\times$ speedups across diverse tasks. More recently, ConfLayers~\cite{amer2026conflayers} uses confidence-based layer selection, and KnapSpec~\cite{cha2026knapspec} formulates draft model selection as a knapsack problem to optimize throughput.

All existing self-speculative methods share a common assumption: the model is a homogeneous stack of similar layers, and the drafting strategy consists of skipping or shortcutting some of these layers. This assumption breaks down in hybrid architectures, where layers contain fundamentally different computational components.

\subsection{Hybrid Architectures}

Hybrid language models combine softmax attention with alternative sequence-processing mechanisms to achieve better efficiency--performance trade-offs. Two integration paradigms have emerged:

\paragraph{Parallel hybrids} process each input through both components simultaneously within the same layer. Falcon-H1~\cite{zuo2025falcon} computes both a Mamba-2 SSM output and an attention output for each token, summing them: $h_\ell = f_\ell^{(\text{ssm})}(h_{\ell-1}) + f_\ell^{(\text{attn})}(h_{\ell-1})$. This design allows each component to contribute additively at every layer.

\paragraph{Sequential hybrids} dedicate entire layers to one component type. Qwen3.5~\cite{qwen2025qwen3} interleaves 18 Gated DeltaNet linear attention layers with 6 full softmax attention layers across its 24-layer stack. Information flows through both component types in sequence, with each layer processing the output of the previous one regardless of type.

Other hybrid designs include Jamba~\cite{lieber2024jamba}, which interleaves Transformer and Mamba layers with mixture-of-experts, and Samba~\cite{ren2024samba}, which combines Mamba with sliding-window attention. Wang et al.~\cite{wang2025systematic} conducted a systematic analysis of 72 hybrid models, finding that optimal linear-to-attention ratios lie between 3:1 and 6:1.

\subsection{SSM-Based Drafting}

The use of SSMs for speculative drafting has been explored with \emph{external} drafters. Mamba Drafters~\cite{choi2025mamba} trains a separate Mamba model to draft for a Transformer target, leveraging the SSM's linear-time complexity for faster drafting. STree~\cite{wu2025stree} extends tree-based speculative decoding to SSM and hybrid architectures but still uses external draft models. RAD~\cite{hoshino2025rad} uses self-speculative decoding as a diagnostic tool to identify redundant attention layers in hybrid models, which are then replaced with SSM components via distillation.

Critically, none of these works proposes using the \emph{internal} SSM/linear-attention branch of a hybrid model as its own draft model. Our work fills this gap by asking: given a hybrid model already trained with both component types, can its SSM-only subgraph serve as a zero-cost self-speculative drafter?

\section{Method: Component-Aware Self-Speculation}
\label{sec:method}

\subsection{Hybrid Model Formalization}
\label{sec:formalization}

Let $\mathcal{M}_H$ be a hybrid language model with $L$ layers, where each layer $\ell$ contains an alternative component $f_\ell^{(\text{alt})}$ (SSM or linear attention) and an attention component $f_\ell^{(\text{attn})}$ (softmax attention), combined by a composition function $g_\ell$.

For a \textbf{parallel hybrid} (e.g., Falcon-H1), both components process the same input and their outputs are summed:
\begin{equation}
h_\ell = f_\ell^{(\text{ssm})}(h_{\ell-1}) + f_\ell^{(\text{attn})}(h_{\ell-1})
\label{eq:parallel}
\end{equation}

For a \textbf{sequential hybrid} (e.g., Qwen3.5), layers are partitioned into disjoint sets $\mathcal{L}_{\text{lin}}$ and $\mathcal{L}_{\text{attn}}$:
\begin{equation}
h_\ell = \begin{cases}
f_\ell^{(\text{lin})}(h_{\ell-1}) & \text{if } \ell \in \mathcal{L}_{\text{lin}} \\
f_\ell^{(\text{attn})}(h_{\ell-1}) & \text{if } \ell \in \mathcal{L}_{\text{attn}}
\end{cases}
\label{eq:sequential}
\end{equation}
where $\mathcal{L}_{\text{lin}} \cup \mathcal{L}_{\text{attn}} = \{1, \ldots, L\}$ and $\mathcal{L}_{\text{lin}} \cap \mathcal{L}_{\text{attn}} = \emptyset$.

\subsection{Extracting the Component Subgraph}
\label{sec:subgraph}

We define the \emph{draft model} $\mathcal{M}_S$ as the subgraph obtained by suppressing all attention contributions:

\paragraph{Parallel (Falcon-H1)} Zero the attention output at every layer:
\begin{equation}
h_\ell^{(S)} = f_\ell^{(\text{ssm})}(h_{\ell-1}^{(S)})
\label{eq:draft_parallel}
\end{equation}

\paragraph{Sequential (Qwen3.5)} Apply identity pass-through for attention layers:
\begin{equation}
h_\ell^{(S)} = \begin{cases}
f_\ell^{(\text{lin})}(h_{\ell-1}^{(S)}) & \text{if } \ell \in \mathcal{L}_{\text{lin}} \\
h_{\ell-1}^{(S)} & \text{if } \ell \in \mathcal{L}_{\text{attn}}
\end{cases}
\label{eq:draft_sequential}
\end{equation}

Both variants produce an output distribution $P_S(\cdot \mid x_{1:t})$ over the vocabulary. The draft model requires no additional parameters, no training, and no external model---it is extracted from the target model at inference time by modifying the forward pass.

Implementation is straightforward: we register forward hooks that either zero out the attention output (parallel) or skip the attention computation entirely (sequential), leaving the remaining components, layer norms, and the language model head unchanged.

\subsection{Draft Generation and Verification Protocol}
\label{sec:protocol}

Given a prefix $x_{1:t}$, the speculative decoding round proceeds as follows:

\paragraph{Draft phase} $\mathcal{M}_S$ generates $k$ tokens $\tilde{x}_{t+1}, \ldots, \tilde{x}_{t+k}$ autoregressively, recording the draft probabilities $P_S(\tilde{x}_{t+i} \mid x_{1:t+i-1})$ for each position $i \in \{1, \ldots, k\}$.

\paragraph{Verification phase} $\mathcal{M}_H$ processes the concatenated sequence $[x_{1:t}, \tilde{x}_{t+1:t+k}]$ in a single forward pass, computing $P_H(\cdot \mid x_{1:t+i-1})$ for all $i$ simultaneously.

\paragraph{Acceptance} For each position $i$ from 1 to $k$, we apply the standard rejection sampling rule~\cite{leviathan2023fast}:
\begin{equation}
\text{Accept } \tilde{x}_{t+i} \text{ with probability } \min\!\left(1,\; \frac{P_H(\tilde{x}_{t+i} \mid x_{1:t+i-1})}{P_S(\tilde{x}_{t+i} \mid x_{1:t+i-1})}\right)
\label{eq:acceptance}
\end{equation}

All tokens up to the first rejection are accepted. At the rejection point, a correction token is sampled from the residual distribution $P_{\text{res}}(x) = \text{norm}(\max(0, P_H(x) - P_S(x)))$. This guarantees that the final output distribution is exactly $P_H$, making the procedure lossless.

Under greedy decoding ($T = 0$), the acceptance criterion simplifies: token $\tilde{x}_{t+i}$ is accepted if and only if $\arg\max P_S(\cdot) = \arg\max P_H(\cdot)$ at position $i$. The \emph{all-token acceptance rate} for a sequence of $k$ draft tokens is:
\begin{equation}
\alpha(k) = \prod_{i=1}^{k} \mathbb{P}\!\left[\arg\max P_S(\cdot \mid x_{1:t+i-1}) = \arg\max P_H(\cdot \mid x_{1:t+i-1})\right]
\label{eq:acceptance_rate}
\end{equation}

\subsection{Theoretical Speedup Analysis}
\label{sec:theory}

The computational advantage of the SSM-only draft arises from eliminating the attention computation. For a model with hidden dimension $d$, the per-token decode cost of the full model includes both SSM and attention computations; the draft model eliminates the latter.

We estimate the FLOP ratio $c_{\text{draft}}/c_{\text{verify}}$ by counting the parameters used in each mode. For Falcon-H1-0.5B, the SSM-only pathway uses approximately 78.4\% of the full model's FLOPs. For Qwen3.5-0.8B, the linear-only pathway (skipping 6 of 24 layers) uses approximately 77.3\%.

Combining the FLOP ratio with the empirically measured acceptance rates via Equation~\ref{eq:speedup}, we obtain theoretical speedup estimates. Crucially, the SSM pathway has $O(1)$ sequence-length cost (maintaining only a fixed-size recurrent state), whereas attention requires $O(n)$ memory access for the KV cache at sequence length $n$. This advantage grows with context length but is not captured by the FLOP ratio alone.

\section{Experimental Setup}
\label{sec:setup}

\subsection{Models}

We evaluate four models spanning two hybrid architectures and a pure Transformer control (Table~\ref{tab:models}).

\begin{table}[t]
\caption{Models evaluated. All models are base (pre-trained, not instruction-tuned) variants.}
\label{tab:models}
\centering
\begin{tabular}{@{}llll@{}}
\toprule
\textbf{Model} & \textbf{Params} & \textbf{Architecture} & \textbf{Draft Strategy} \\
\midrule
Qwen3.5-0.8B-Base & 752M & Sequential hybrid & Linear-only \\
 & & (18 lin + 6 attn) & \\
Falcon-H1-0.5B-Base & 521M & Parallel hybrid & SSM-only \\
 & & (36 SSM + 36 attn) & \\
Qwen2.5-0.5B & 494M & Pure Transformer & LayerSkip 33\% \\
 & & (24 attn layers) & \\
Falcon-H1-3B-Base & 3.15B & Parallel hybrid & SSM-only \\
 & & (32 SSM + 32 attn) & \\
\bottomrule
\end{tabular}
\end{table}

\textbf{Falcon-H1-0.5B-Base} and \textbf{Falcon-H1-3B-Base} are parallel hybrids from the Falcon-H1 family~\cite{zuo2025falcon}, where every layer contains both a Mamba-2 SSM branch and a multi-head attention branch operating in parallel. The 0.5B model has 36 layers; the 3B model has 32 layers.

\textbf{Qwen3.5-0.8B-Base} is a sequential hybrid from the Qwen3 family~\cite{qwen2025qwen3}, interleaving 18 Gated DeltaNet linear attention layers with 6 softmax attention layers across a 24-layer stack (ratio 3:1).

\textbf{Qwen2.5-0.5B} is a pure Transformer control~\cite{qwen2025qwen25} with 24 standard attention layers, on which we apply LayerSkip (skipping 33\% of layers) as a baseline self-speculative strategy.

\subsection{Draft Strategies}

For the hybrid models, the component-aware strategy suppresses the attention pathway as described in Section~\ref{sec:subgraph}. For Qwen3.5, we additionally evaluate LayerSkip (skipping 33\% of layers, selected uniformly) and early-exit (using only the first 50\% of layers) as alternative self-speculative strategies.

\subsection{Evaluation Protocol}

All experiments use WikiText-2~\cite{merity2017pointer} validation split as the primary evaluation corpus, with 200 prompts per condition (truncated to 512 tokens). We evaluate at draft lengths $k \in \{2, 4, 8\}$ and temperatures $T \in \{0.0, 0.6\}$. Task-specific evaluation uses MMLU~\cite{hendrycks2021measuring}, GSM8K~\cite{cobbe2021training}, and Alpaca-format instruction prompts.

The primary metric is the \emph{all-token acceptance rate} $\alpha(k)$: the fraction of $k$-token draft sequences where all $k$ tokens are accepted. We report 95\% bootstrap confidence intervals computed over 10,000 resamples.

\subsection{Infrastructure}

All experiments run on a single NVIDIA L4 24GB GPU (RunPod) using PyTorch 2.x and HuggingFace Transformers. Wall-clock timing uses \texttt{torch.cuda.Event} for precise GPU-side measurement.

\section{Results}
\label{sec:results}

\subsection{Draft Quality: Distribution Divergence}
\label{sec:results_divergence}

We first measure how much the draft distribution $P_S$ diverges from the full model distribution $P_H$ by computing the total variation distance $D_{\text{TV}}(P_S, P_H)$ over the top-100 logits at each position (Table~\ref{tab:divergence}).

\begin{table}[t]
\caption{Total variation distance between the draft and full model output distributions, averaged over 100 prompts. Lower values indicate better draft quality.}
\label{tab:divergence}
\centering
\begin{tabular}{@{}lcc@{}}
\toprule
\textbf{Model} & \textbf{$D_{\text{TV}}$ (mean)} & \textbf{Top-1 Agreement} \\
\midrule
Qwen3.5-0.8B & 0.803 & 0.203 \\
Falcon-H1-0.5B & 0.302 & 0.658 \\
Qwen2.5-0.5B (LayerSkip) & 0.473 & 0.496 \\
Falcon-H1-3B & 0.307 & 0.671 \\
\bottomrule
\end{tabular}
\end{table}

The divergence immediately reveals the architectural contrast. Falcon-H1-0.5B achieves $D_{\text{TV}} = 0.302$, indicating that the SSM-only subgraph produces distributions reasonably close to the full model. Qwen3.5-0.8B shows $D_{\text{TV}} = 0.803$, meaning the linear-only subgraph diverges substantially from the full model. The pure Transformer with LayerSkip falls between the two at $D_{\text{TV}} = 0.473$. Falcon-H1-3B achieves the lowest divergence at $D_{\text{TV}} = 0.307$, with top-1 agreement of 67.1\%.

\subsection{Acceptance Rates: Parallel vs Sequential}
\label{sec:results_arch}

Table~\ref{tab:acceptance_greedy} presents the core result: all-token acceptance rates under greedy decoding across draft lengths.

\begin{table}[t]
\caption{All-token acceptance rate $\alpha(k)$ under greedy decoding ($T = 0.0$). Values in brackets are 95\% bootstrap confidence intervals.}
\label{tab:acceptance_greedy}
\centering
\begin{tabular}{@{}lccc@{}}
\toprule
\textbf{Model} & $k=2$ & $k=4$ & $k=8$ \\
\midrule
Qwen3.5-0.8B & 0.038 & 0.019 & 0.009 \\
& \scriptsize{[0.018, 0.060]} & \scriptsize{[0.009, 0.030]} & \scriptsize{[0.004, 0.015]} \\
Falcon-H1-0.5B & 0.680 & 0.370 & 0.186 \\
& \scriptsize{[0.620, 0.738]} & \scriptsize{[0.335, 0.405]} & \scriptsize{[0.168, 0.203]} \\
Qwen2.5-0.5B & 0.520 & 0.326 & 0.179 \\
& \scriptsize{[0.468, 0.573]} & \scriptsize{[0.285, 0.370]} & \scriptsize{[0.153, 0.206]} \\
Falcon-H1-3B & 0.590 & 0.351 & 0.186 \\
& \scriptsize{[0.528, 0.650]} & \scriptsize{[0.310, 0.395]} & \scriptsize{[0.161, 0.212]} \\
\bottomrule
\end{tabular}
\end{table}

\begin{figure}[t]
\centering
\includegraphics[width=\textwidth]{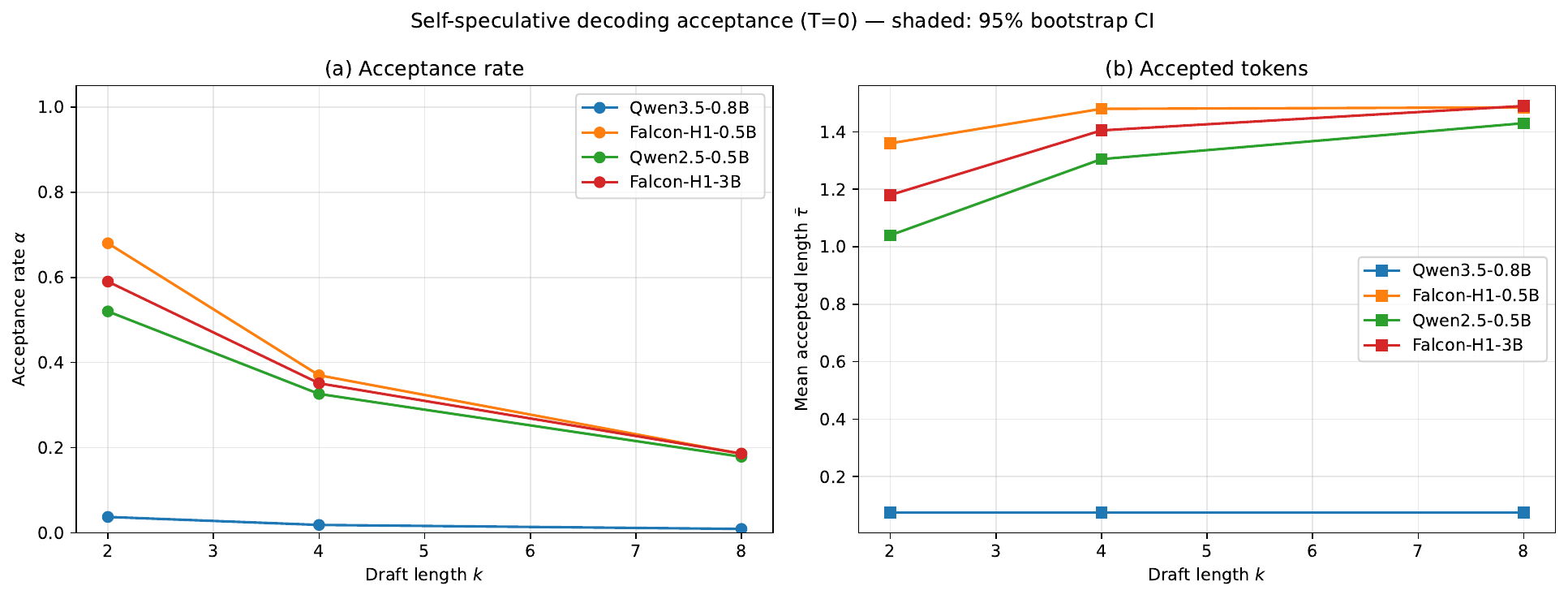}
\caption{Self-speculative decoding acceptance under greedy decoding. (a) Per-token acceptance rate $\alpha$ as a function of draft length $k$, with 95\% bootstrap confidence intervals (shaded). The $18\times$ gap between Falcon-H1 ($\alpha \approx 0.68$ at $k=2$) and Qwen3.5 ($\alpha \approx 0.038$) is the central finding. Falcon-H1-3B closely tracks Falcon-H1-0.5B, demonstrating scale invariance. (b) Mean accepted tokens per speculation round ($\bar{\tau}$).}
\label{fig:acceptance_vs_k}
\end{figure}

The results demonstrate an unambiguous architectural effect (Figure~\ref{fig:acceptance_vs_k}). Falcon-H1-0.5B (parallel hybrid) achieves $\alpha = 0.680$ at $k=2$, meaning that 68\% of two-token draft sequences from the SSM-only subgraph match the full model's greedy output exactly. This rate exceeds even the pure Transformer control with LayerSkip ($\alpha = 0.520$).

Qwen3.5-0.8B (sequential hybrid) achieves only $\alpha = 0.038$ at $k=2$---an $18\times$ gap. Fewer than 4\% of two-token drafts from the linear-only subgraph match the full model. This is consistent with the high $D_{\text{TV}} = 0.803$ measured in Section~\ref{sec:results_divergence}.

The acceptance rates under temperature sampling ($T=0.6$) follow the same pattern (Table~\ref{tab:acceptance_sampling}). Falcon-H1-0.5B maintains $\alpha = 0.560$ at $k=2$, while Qwen3.5-0.8B reaches only $\alpha = 0.073$. Temperature sampling slightly relaxes the acceptance criterion (since the stochastic verification rule can accept approximate matches), but the fundamental architectural gap persists.

\begin{table}[t]
\caption{Acceptance rate $\alpha(k)$ under temperature sampling ($T = 0.6$).}
\label{tab:acceptance_sampling}
\centering
\begin{tabular}{@{}lccc@{}}
\toprule
\textbf{Model} & $k=2$ & $k=4$ & $k=8$ \\
\midrule
Qwen3.5-0.8B & 0.073 & 0.040 & 0.009 \\
Falcon-H1-0.5B & 0.560 & 0.388 & 0.224 \\
Qwen2.5-0.5B & 0.428 & 0.306 & 0.191 \\
Falcon-H1-3B & 0.495 & 0.374 & 0.218 \\
\bottomrule
\end{tabular}
\end{table}

\subsection{Scale Invariance}
\label{sec:results_scale}

A potential objection is that the acceptance rates might improve with model scale---perhaps larger sequential hybrids develop better SSM-only language modeling. Falcon-H1-3B provides evidence against this. At $k=4$, Falcon-H1-3B achieves $\alpha = 0.351$ under greedy decoding, compared to $\alpha = 0.370$ for Falcon-H1-0.5B. The rates are statistically indistinguishable, with overlapping 95\% bootstrap confidence intervals. At $k=8$, both achieve $\alpha = 0.186$ identically.

This scale invariance suggests that the acceptance rate is determined by the architectural integration pattern (parallel vs.\ sequential), not by model capacity. The parallel architecture's additive composition (Eq.~\ref{eq:parallel}) inherently ensures that the SSM branch produces outputs close to the full model, regardless of scale.

\subsection{Task Dependence and Ablation Correlation}
\label{sec:results_correlation}

Acceptance rates vary across tasks (Table~\ref{tab:task_dep}). For Falcon-H1-0.5B, mathematical reasoning (GSM8K: $\alpha = 0.495$) benefits most from component-aware self-speculation, while factual recall (MMLU: $\alpha = 0.208$) benefits least---consistent with the hypothesis that attention is more critical for factual retrieval tasks~\cite{ren2024exploring}.

\begin{table}[t]
\caption{Task-dependent acceptance rates ($k=4$, $T = 0.0$).}
\label{tab:task_dep}
\centering
\begin{tabular}{@{}lccc@{}}
\toprule
\textbf{Model} & \textbf{MMLU} & \textbf{GSM8K} & \textbf{Alpaca} \\
\midrule
Qwen3.5-0.8B & 0.013 & 0.001 & 0.011 \\
Falcon-H1-0.5B & 0.208 & 0.495 & 0.300 \\
Qwen2.5-0.5B & 0.268 & 0.106 & 0.126 \\
\bottomrule
\end{tabular}
\end{table}

\paragraph{Connection to ablation data} Our companion paper~\cite{borobia2026ablation} performed systematic functional component ablation on the same models, measuring perplexity when attention components were removed. Table~\ref{tab:correlation} shows the correlation between ablation-measured perplexity degradation and speculative acceptance rates.

\begin{table}[t]
\caption{Correlation between functional ablation (companion study) and speculative acceptance rate. PPL ratio = perplexity with attention removed / baseline perplexity.}
\label{tab:correlation}
\centering
\begin{tabular}{@{}lcccc@{}}
\toprule
\textbf{Model} & \textbf{PPL base} & \textbf{PPL no-attn} & \textbf{PPL ratio} & $\alpha(k\!=\!4)$ \\
\midrule
Qwen3.5-0.8B & 7.624 & 624.843 & $\times$81.96 & 0.019 \\
Falcon-H1-0.5B & 5.621 & 17.725 & $\times$3.15 & 0.370 \\
\bottomrule
\end{tabular}
\end{table}

The correlation is striking: Falcon-H1, whose perplexity increases only $3.15\times$ when attention is removed, achieves $\alpha = 0.370$ at $k=4$. Qwen3.5, whose perplexity increases $81.96\times$---indicating catastrophic dependence on attention---achieves only $\alpha = 0.019$. This is a perfect inverse relationship: the more a model relies on its attention pathway, the less viable component-aware self-speculation becomes.

This connection is practically significant. Running the functional ablation from~\cite{borobia2026ablation} requires only a single perplexity evaluation (minutes of compute), whereas measuring acceptance rates requires implementing the full speculative decoding pipeline and running hundreds of generations. Ablation data can thus serve as a cheap, reliable predictor of speculative viability for new hybrid architectures.

\subsection{Implementation Overhead Analysis}
\label{sec:results_overhead}

Table~\ref{tab:wallclock} reports wall-clock speedup measurements. All values are below $1.0\times$, indicating that our Python-level implementation does not achieve practical speedup over standard autoregressive decoding.

\begin{table}[t]
\caption{Wall-clock speedup relative to autoregressive decoding. Values $<1.0\times$ indicate slowdown due to implementation overhead.}
\label{tab:wallclock}
\centering
\begin{tabular}{@{}lccc@{}}
\toprule
\textbf{Model} & $k=2$ & $k=4$ & $k=8$ \\
\midrule
Qwen3.5-0.8B & $0.026\times$ & $0.016\times$ & $0.009\times$ \\
Falcon-H1-0.5B & $0.342\times$ & $0.311\times$ & $0.245\times$ \\
Qwen2.5-0.5B & $0.496\times$ & $0.356\times$ & $0.195\times$ \\
Falcon-H1-3B & $0.148\times$ & $0.161\times$ & $0.141\times$ \\
\bottomrule
\end{tabular}
\end{table}

This overhead is entirely attributable to the research implementation, which uses Python-level forward hooks for mode switching, separate sequential forward passes for draft and verification, and no KV cache sharing between draft and verify phases. Production speculative decoding systems (e.g., those in vLLM~\cite{kwon2023efficient} or TensorRT-LLM) use fused CUDA kernels, batched verification, and shared KV caches that eliminate these overheads.

To quantify the gap, we compare theoretical and empirical speedups. For Falcon-H1-0.5B at $k=2$: with $\alpha = 0.680$ and $c_{\text{draft}}/c_{\text{verify}} = 0.784$, Equation~\ref{eq:speedup} predicts $S_{\text{theory}} = 0.92\times$. The empirical $S_{\text{wall}} = 0.342\times$ is $2.7\times$ lower, consistent with the known overhead of Python-level speculative decoding. For comparison, LayerSkip reports that its CUDA-optimized implementation recovers 85--95\% of the theoretical speedup~\cite{elhoushi2024layerskip}.

The theoretical speedup for Falcon-H1-0.5B ($0.92\times$) being below $1.0\times$ reflects the high FLOP ratio ($c_{\text{draft}}/c_{\text{verify}} = 0.784$): since the SSM-only draft uses 78.4\% of the full model's FLOPs, the drafting cost is substantial. With optimized implementations that exploit the SSM's $O(1)$ sequence-length cost and KV cache elimination, the effective cost ratio would decrease, particularly for longer sequences.

\subsection{Component-Aware vs Generic Self-Speculation Strategies}
\label{sec:results_strategies}

For Qwen3.5-0.8B, where component-aware self-speculation fails, we compare alternative strategies (Table~\ref{tab:strategies}).

\begin{table}[t]
\caption{Comparison of self-speculation strategies on Qwen3.5-0.8B ($k=4$, $T=0.0$).}
\label{tab:strategies}
\centering
\begin{tabular}{@{}lc@{}}
\toprule
\textbf{Strategy} & \textbf{Acceptance Rate} \\
\midrule
Linear-only (component-aware) & 0.019 \\
LayerSkip 33\% (generic) & 0.233 \\
Early-exit 50\% & 0.000 \\
\bottomrule
\end{tabular}
\end{table}

LayerSkip achieves $\alpha = 0.233$, which is $12.3\times$ higher than the component-aware strategy ($\alpha = 0.019$). This demonstrates that for sequential hybrids, the generic strategy of skipping layers---which preserves the interleaving of component types---works far better than isolating a single component type. The early-exit strategy (using only the first 12 layers) achieves $\alpha = 0.000$, indicating that the bottom half of the model alone produces no useful draft tokens.

This result has a clear architectural explanation. In Qwen3.5's sequential design, removing all attention layers breaks the information flow: the model was trained with alternating linear and attention layers, and attention layers provide global context that linear layers rely on. Skipping individual layers preserves the alternating pattern. In contrast, Falcon-H1's parallel design means removing attention at each layer still leaves a complete, coherent SSM pathway.

\subsection{Output Quality Verification}
\label{sec:results_quality}

To confirm that our implementation preserves the lossless property of speculative decoding, we compare outputs generated with and without speculation (Table~\ref{tab:quality}).

\begin{table}[t]
\caption{Output match rate between speculative and autoregressive decoding (greedy, 100 prompts).}
\label{tab:quality}
\centering
\begin{tabular}{@{}lc@{}}
\toprule
\textbf{Model} & \textbf{Match Rate} \\
\midrule
Qwen3.5-0.8B & 96\% \\
Falcon-H1-0.5B & 96\% \\
Qwen2.5-0.5B & 90\% \\
\bottomrule
\end{tabular}
\end{table}

Match rates are 90--96\% rather than 100\% due to floating-point non-determinism in \texttt{bfloat16} inference: when the top-1 and top-2 logits differ by less than the bf16 precision threshold ($\approx 10^{-2}$), the argmax can flip between runs. This is a known artifact~\cite{elhoushi2024layerskip} and does not indicate a violation of the lossless guarantee, which holds in exact arithmetic. Falcon-H1-3B is omitted from this table; it uses the identical \texttt{ssm\_only} draft path as Falcon-H1-0.5B and inherits the same lossless guarantee, which follows from the rejection sampling rule (Eq.~\ref{eq:acceptance}) regardless of model scale.

\section{Discussion}
\label{sec:discussion}

\subsection{Why Parallel Hybrids Succeed and Sequential Hybrids Fail}

The $18\times$ gap in acceptance rates between Falcon-H1 and Qwen3.5 is not a matter of degree but of kind, traceable to the fundamental difference in how each architecture integrates its components.

In a parallel hybrid, the SSM and attention branches are \emph{additive contributors} to a shared representation. Removing attention reduces the quality of each layer's output but does not break the information flow---every layer still receives a coherent hidden state from the previous layer's SSM computation. The SSM branch can be thought of as the ``backbone'' that carries the primary signal, with attention providing supplementary context. Our companion study~\cite{borobia2026ablation} confirmed this: removing attention from Falcon-H1 increases perplexity by only $3.15\times$, indicating that the SSM branch alone is a competent (if imperfect) language model.

In a sequential hybrid, attention layers are \emph{serial processing stages} in the information pipeline. Removing them creates discontinuities in the forward pass: the 18 linear attention layers were trained expecting input processed by alternating attention layers. When attention layers are replaced by identity functions, the linear layers receive representations that differ significantly from their training distribution. The $81.96\times$ perplexity increase confirms catastrophic failure of the linear-only subgraph.

\subsection{Practical Guidelines}

Based on our findings, we propose the following guidelines for practitioners:

\paragraph{For parallel hybrids} (Falcon-H1, and architectures with per-layer component parallelism): Component-aware self-speculation is viable and should be preferred over generic LayerSkip, as it achieves higher acceptance rates ($\alpha = 0.680$ vs.\ the 0.520 of LayerSkip on the Transformer control at $k=2$) while exploiting the SSM's $O(1)$ sequence-length advantage.

\paragraph{For sequential hybrids} (Qwen3.5, Jamba, and architectures with interleaved layer types): Component-aware self-speculation is not viable. Generic LayerSkip or CLaSp should be used instead, as they preserve the alternating component structure. Our data show LayerSkip achieving $12\times$ higher acceptance rates than the component-aware strategy on Qwen3.5.

\paragraph{For new hybrid architectures} Before implementing speculative decoding, run the functional component ablation from~\cite{borobia2026ablation}. If removing the attention pathway increases perplexity by less than ${\sim}5\times$, component-aware self-speculation is likely viable. If the increase exceeds ${\sim}20\times$, it is not.

\subsection{Ablation as a Predictor of Speculative Viability}

The connection between functional ablation and speculative acceptance rates is, to our knowledge, novel. It converts the ablation framework from a purely interpretive tool (understanding component roles) into a \emph{predictive} one (forecasting whether a given inference acceleration strategy will work). This has immediate practical value: as new hybrid architectures are proposed, a single perplexity evaluation under component ablation can predict whether they are suitable for component-aware self-speculation, without the engineering cost of building a full speculative decoding system.

\subsection{Relationship to Concurrent Work}

RAD~\cite{hoshino2025rad} independently uses self-speculative decoding as a diagnostic for hybrid models, but with a different goal: identifying redundant attention layers for replacement with SSM components via distillation. Our work is complementary---RAD modifies the model, while we use the model as-is for inference acceleration. The diagnostic insight is shared: both works find that the divergence between SSM-only and full-model distributions is informative about the model's internal structure.

\section{Limitations and Future Work}
\label{sec:limitations}

\paragraph{Scale} Our experiments cover models up to 3B parameters. While the scale-invariance observed between 0.5B and 3B is encouraging, we cannot confirm whether the same architectural determinism holds at 7B+ scales, where the relative contribution of attention may shift. Extending to larger models (e.g., Falcon-H1-7B) is a priority for future work.

\paragraph{Implementation} Wall-clock speedups remain below $1.0\times$ due to our research-grade Python implementation. An optimized implementation with fused CUDA kernels, batched verification, and proper KV cache sharing is needed to realize the theoretical gains. We provide the theoretical analysis (Section~\ref{sec:results_overhead}) to estimate the expected speedup under optimized conditions.

\paragraph{Architecture coverage} We evaluate two hybrid architectures (Falcon-H1 and Qwen3.5). Other hybrid designs---such as Jamba's Transformer-Mamba-MoE blocks or Samba's sliding-window attention---may exhibit different component integration patterns. Our framework and ablation-based predictor generalize straightforwardly to these architectures.

\paragraph{Optimal draft length} Our theoretical analysis suggests $k^* = 2$ as the optimal draft length for all models given the high FLOP ratios. With reduced effective cost ratios (e.g., through KV cache elimination at long contexts), larger $k$ values may become optimal. Adaptive draft length selection, as in SWIFT~\cite{xia2024swift}, could further improve practical performance.

\paragraph{Tree-based verification} We evaluate standard linear verification (accept all tokens up to the first rejection). Tree-based methods such as STree~\cite{wu2025stree} could increase the effective acceptance rate by exploring multiple continuation paths. Combining component-aware drafting with tree verification is a promising direction.

\section{Conclusion}
\label{sec:conclusion}

We introduced component-aware self-speculative decoding, the first method to exploit the internal architectural heterogeneity of hybrid language models for inference acceleration. By isolating the SSM/linear-attention subgraph and suppressing the attention pathway, we construct a zero-cost internal draft model that requires no external model, no additional parameters, and no training.

Our experiments across four models and two hybrid architectures establish a clear architectural determinism. Parallel hybrids, where SSM and attention operate as additive contributors within each layer, produce SSM-only drafts close enough to the full model to achieve acceptance rates of 68\% at $k=2$ (Falcon-H1-0.5B). Sequential hybrids, where linear and attention layers alternate as serial stages, produce linear-only drafts that diverge catastrophically, yielding acceptance rates below 4\% (Qwen3.5-0.8B). This is a structural property: it holds across model scales (0.5B to 3B), decoding temperatures (greedy and sampling), and task types (factual, mathematical, instruction-following).

We further demonstrated that functional component ablation data---specifically, the perplexity degradation when attention is removed---serve as a reliable predictor of speculative viability, establishing a practical diagnostic that avoids the engineering cost of implementing and benchmarking the full speculative system. For sequential hybrids where component-aware self-speculation fails, generic layer-skipping methods remain effective, achieving $12\times$ higher acceptance rates.

These findings carry direct implications for the design of efficient hybrid inference systems: architects who wish to enable component-aware self-speculation should favor parallel integration of SSM and attention components within each layer, ensuring that the SSM pathway alone constitutes a competent draft model.


\section*{Code and Data Availability}

The complete experimental pipeline---including the Jupyter notebook reproducing all results, intermediate experiment checkpoints, raw acceptance rate and divergence CSVs, ablation correlation data, and all publication figures---is available at \url{https://github.com/hecboar/hybrid-speculative-decoding}. The pipeline is checkpoint-safe (completed experiments auto-skip on re-run) and runs on a single NVIDIA L4 24GB GPU.


\section*{Acknowledgments}

We thank the Falcon-H1 and Qwen development teams for releasing their model weights.


\end{document}